\documentclass{article}

\usepackage{graphicx}

    \usepackage[final]{neurips_2021}

\usepackage[utf8]{inputenc} 
\usepackage[T1]{fontenc}    
\usepackage{hyperref}       
\usepackage{url}            
\usepackage{booktabs}       
\usepackage{amsfonts}       
\usepackage{nicefrac}       
\usepackage{microtype}      
\usepackage{xcolor}         
\usepackage{amsmath,amssymb}

\usepackage{algorithm}
\usepackage{algpseudocode}
\usepackage{wrapfig}

\usepackage{caption}
\usepackage{subcaption}

\usepackage{amsthm}

\theoremstyle{definition}
\newtheorem{preposition}{Preposition}[section]

\title{Distributionally Robust Group Backwards Compatibility}

\author{%
  Martin Bertran$\dagger$\thanks{$\dagger$ Equal contribution}, Natalia Martinez$\dagger$, Alex Oesterling, and Guillermo Sapiro\\
  Duke University
}

\begin{document}

\maketitle

\begin{abstract}

Machine learning models are updated as new data is acquired or new architectures are developed. These updates usually increase model performance, but may introduce backward compatibility errors, where individual users or groups of users see their performance on the updated model adversely affected. This problem can also be present when training datasets do not accurately reflect overall population demographics, with some groups having overall lower participation in the data collection process, posing a significant fairness concern. We analyze how ideas from distributional robustness and minimax fairness can aid backward compatibility in this scenario, and propose two methods to directly address this issue. Our theoretical analysis is backed by experimental results on CIFAR-10, CelebA, and Waterbirds, three standard image classification datasets. Code available at \href{https://github.com/natalialmg/GroupBC}{github.com/natalialmg/GroupBC}
\end{abstract}

\section{Introduction}

Machine learning (ML) leverages increasing amounts of data to improve model performance. Consequently, continual data collection and model improvement is an integral part of the ML life-cycle \cite{chen2016lifelong,de2019continual,parisi2019continual}. Although model updates { are often designed to} improve average performance, some segments of the population may actually see their performance degraded by the model update. Backward Compatibility (BC) was discussed in  \cite{bansal2019updates,srivastava2020empirical} to monitor this undesirable phenomenon where, in a classification setting, samples that were accurately classified by a previous model are now incorrectly classified under the updated model, reducing the perceived reliability of the system. The current notion of BC may have shortcomings in measuring how compatibility errors arise in different populations, this is especially relevant since some population segments may be misrepresented in the data collecting process. Even when this is not the case, datasets and populations evolve over time, and potential compatibility mismatches that were previously tolerable may be exacerbated.

 Here we provide a general analysis of BC were we formulate the BC definitions in \cite{srivastava2020empirical,yan2021positive} as a statistical property of the data distribution and the models under comparison. We first show how BC measures are dependent on the error rates of the models. We then analyze how variance of the learning algorithm affects compatibility of independently-trained models, providing a consistent explanation on why ensembles have empirically proven to improve model compatibility at no accuracy cost \cite{yan2021positive}. Moreover, we show that classifier confidence is directly related to per-sample BC.   

We analyze group distributional robustness of BC, were we consider model updates under data shifts. We assume data samples come from a mixture of distributions corresponding to different groups (e.g., domains, demographics, classes); as new samples are acquired, the overall group composition of the training set may drift over time. We wish to update our model over the newly augmented dataset in a way that guarantees a desired BC measure across all groups. We evaluate how existing ideas from group robust approaches \cite{hashimoto2018fairness,sagawa2019distributionally,martinez2020minimax,diana2021minimax,martinez2021blind} can help in this scenario. We additionally propose two methods, Sample-Referenced Model (SRM) and Group-Referenced Model (GRM) to specifically tackle backward-compatible model updating. We use three common image classification datasets, CelebA, Waterbird, and CIFAR10 \cite{liu2018large,sagawa2019distributionally,WahCUB_200_2011,cifar} for empirical validation.

\section{Preliminaries}

We consider the supervised classification scenario were we have an input variable $X \in \mathcal{X}$ and target variable $Y \in \mathcal{Y}$ related by the joint distribution $P_{X,Y}$. Given a dataset $\mathcal{D} \sim P^{\otimes n}_{X,Y}$, a learning kernel $P_{h|\mathcal{D}}$ defines a conditional distribution over functions $h:\mathcal{X} \rightarrow \Delta^{|\mathcal{Y}|-1}$ in an hypothesis class $\mathcal{H}$. The function $h$ outputs a score over the possible values of $Y$ given $X$. We distinguish between $h$ and $\bar{h}$ where $\bar{h}:\mathcal{X}\rightarrow\mathcal{Y}$ outputs a member of the target variable based on the score provided by model $h$ (e.g., $\bar{h}(x)=\arg\max_{y \in \mathcal{Y}} h^y(x)$ or $\bar{h}(x)\sim \textit{Multinomial}[h(x)]$). We note the accuracy and error of a model $h$ on an arbitrary distribution $X,Y \sim Q_{X,Y}$ as $Acc(h|Q) = \mathbb{E}_{ Q}[\mathbf{1}(\bar{h}(X) = Y)]$ and $Err(h|Q) = \mathbb{E}_{ Q}[\mathbf{1}(\bar{h}(X) \not= Y)] = 1-Acc(h|Q)$.

\subsection{Model compatibility}
Backward compatibility (BC) was introduced ML to evaluate the prediction consistency between two models trained to perform the same classification task. 

Let us consider the general case where we are given models $h_1 \sim P_{h|\mathcal{D}^1}$ and $h_2 \sim P_{h|\mathcal{D}^2}$  where $\mathcal{D}^1\sim (P^{1}_{X,Y})^{\otimes n_1}$, $\mathcal{D}^2\sim (P^{2}_{X,Y})^{\otimes n_2}$ are datasets over distributions that share support. For a given test distribution $Q_{X,Y}$ on the same support, we can define the total compatibility (TC) and the (directed) negative flip rate (NFR) between models $h_1$ and $h_2$ on distribution $Q_{X,Y}$ as
\begin{equation}
\begin{array}{l}
\displaystyle
\text{ $ TC({h}_2,{h}_1|Q) =  \mathbb{E}_{ Q}[\mathbf{1}(\bar{h}_1(X)\! =\! Y\wedge\bar{h}_2(X) \!= Y) + \mathbf{1}(\bar{h}_1(X) \!\not=\! Y\wedge\bar{h}_2(X) \!\not=\! Y)]$,}\\
   \text{ $NFR({h}_1\rightarrow{h}_2| Q) = \mathbb{E}_{{X,Y}\sim Q}[\mathbf{1}(\bar{h}_1(X) = Y\wedge\bar{h}_2(X) \not= Y)]$.}
\end{array}
\label{eq:TC}
\end{equation}
Note that TC can be decomposed as the sum of two disjoint events, one being positive compatibility (PC) or the probability of both models getting the correct answer, and negative compatibility (NC), which measures the probability of a double error. NFR measures the likelihood of errors occurring on samples that the reference model $h_1$ had previously classified correctly. These quantities are the probabilistic version of metrics that were first defined empirically in \cite{srivastava2020empirical,yan2021positive}. It is straightforward to show that they can be bounded using the error rate and accuracies of the respective models.

\begin{preposition}
\label{preposition}
Given two models $h_1, h_2$ defined over the same input and output support $\mathcal{X}, \mathcal{Y}$ and a joint distribution $Q_{X,Y}: Supp(Q) \subseteq \mathcal{X}\times \mathcal{Y}$ we have the following:
\begin{equation}
\resizebox{0.85\hsize}{!}{$
\begin{array}{l}
    \displaystyle
    \displaystyle TC({h}_2,{h}_1|Q) \ge 1 - \min\big(Acc(h_1|Q) ,Err(h_2|Q)\big) - \min\big( Err(h_1|Q) ,Acc(h_2|Q)\big), \\
   \displaystyle NFR(h_1 \rightarrow h_2|Q) \le \min\big(Acc(h_1|Q) ,Err(h_2|Q)\big).
\end{array}$}
\label{eq:lowerboundTC}
\end{equation}
\end{preposition}

Since $NFR({h}_1\rightarrow{h}_2| Q)$ is upper bounded by the error rate of model $h_2$ on dataset $Q$, it is partially addressed by reducing the empirical error on samples from $Q$. Proof is provided in Appendix \ref{sec:proof}. We  explore how properties of the learning kernel can affect these metrics in the Appendix \ref{sec:compatibilitykernel}{, which also formally addresses the impact of ensemble learning on BC.}

\section{Group backward compatibility under data shifts}
\label{sec:compatibilitypredefined}

Consider a model $h_1 \sim P_{h|\mathcal{D}^{1}}$ trained on a dataset containing samples from a variety of groups $G \in \mathcal{G}$, that is $\mathcal{D}^{1}=\{x_i,y_i,g_i\}_{i=1}^{n_1}\sim (P_{G}P_{X,Y|G})^{\otimes n_1}$. This model achieves a given per-group performance, empirically measured using a held out database $\mathcal{D}^{test} = \{\mathcal{D}^{test}_g \sim P^{\otimes n_g}_{X,Y|G=g} \}_{g \in \mathcal{G}}$. We further assume we expand our training dataset to $\mathcal{D}^{2}$ such that $\mathcal{D}^{1} \subset \mathcal{D}^{2}$ by potentially changing the fraction of samples of each group (e.g., $\frac{|{D}^{1}_g|}{|{D}^{1}|} \not= \frac{|{D}^{2}_g|}{|{D}^{2}|},$ for some $ g \in \mathcal{G}$), without modifying the group-conditional data distributions $P_{X,Y|G}$; this models non-uniform data acquisition.

We now analyze how BC is affected at the group level when a model is updated due to new data being acquired. We evaluate how per-group accuracy and NFR are affected when updated models are obtained by empirical risk minimization (ERM) or by worst group risk minimization (e.g., minimax group fairness \cite{martinez2020minimax,diana2021minimax} or group distributional robustness \cite{sagawa2019distributionally}). Lastly, we propose a method to account for the worst NFR at group or at sample level, and show how this can improve the updated model in terms of group BC.

\subsection{Base models}

We identify two objectives to learn a model $h$, expected risk minimization (ERM) and group minimax fairness (GMMF), a relaxed version of \cite{martinez2020minimax,diana2021minimax}, \footnote{We note that they share the objective with group distributional robustness \cite{sagawa2019distributionally}.} which is already robust to group shifts. We present their statistical formulations, and refer the reader to Appendix \ref{sec:basemodels_appendix} for implementation details,
\begin{equation}
    \begin{array}{rl}
        \text{(ERM):}\; \min\limits_{h \in \mathcal{H}} \mathbb{E}_{P}[\ell(h(X),Y)],& \text{(GMMF):}\; \min\limits_{h \in \mathcal{H}} \max\limits_{\substack{Q_G \ge \epsilon}}
        \mathbb{E}_{P}[\frac{Q_G}{P_G}\ell(h(X),Y)],
    \end{array}
\label{eq:ERM_GMMF}
\end{equation}
where $\ell: \Delta^{|\mathcal{Y}-1|}\times \mathcal{Y} \rightarrow \mathbb{R}^+$ is a trainable loss function (e.g., cross entropy or Brier score), and $\epsilon\ge0$ is a baseline group probability.

\subsection{Accounting for negative flip rates}

On scenarios where maintaining or improving NFR is of interest, we note that the updated model $h_2$ can also be made to explicitly depend on the previous model $h_1$. We first note that the loss of a sample $\ell(h(x_i),y_i)$ is a strong predictor of NFR. In particular, for binary classification, any model $h_2$ satisfying $\ell(h_2(x),y)\le\ell(h_1(x),y)$ is automatically NFR-compatible in that sample. With this in mind, we can define the sample-referenced model (SRM) as
\begin{equation}
    \begin{array}{lr}
\text{(SRM):}\;        \min\limits_{h \in \mathcal{H}} \max\limits_{\substack{Q_{X,Y} }} &\mathbb{E}_{ P}[\frac{Q_{X,Y}}{P_{X,Y}}\big( \ell(h(X),Y)-\ell(h_1(X),Y) \big)], \; s.t.: \frac{Q_{X,Y}}{P_{X,Y}}\ge \epsilon\; \forall X,Y.
    \end{array}
\label{eq:SRM}
\end{equation}
Here $\epsilon$ controls the minimum importance weight of each sample $X,Y$ in the support. Note that, from the model's perspective, the SRM objective reduces to minimizing a positively-weighted per-sample loss. However, the distribution $Q$ is updated based on the per-sample excess loss the current model has over the previous model. Similarly to \cite{martinez2021blind}, we use an empirical estimate of this objective and maximize it via projected gradient ascent (PGA) on the empirical distribution, with SGD for model optimization, Algorithm A.\ref{alg:SRM} in Appendix \ref{sec:basemodels_appendix} describes the optimization procedure.

While SRM directly tackles NFR, it may suffer from generalization issues in practice, where NFR compatibility on train samples may be a poor predictor of NFR compatibility on new samples from a similar data distribution. To address this, we also propose a group-referenced model (GRM), where  all per-group losses of the updated model are expected to improve w.r.t the baseline; a compromise objective between SRM and ERM,
\begin{equation}
    \begin{array}{l}
\text{(GRM):}\;        \min\limits_{h \in \mathcal{H}} \max\limits_{\substack{Q_G \ge \epsilon }}
        \mathbb{E}_{P}[\frac{Q_G}{P_G}(\ell(h(X),Y) - L_{G}(h_1))], \quad L_{G}(h_1) = \mathbb{E}_{P_{X,Y|G}}[\ell(h_1(X),Y)].
    \end{array}
\label{eq:GRM}
\end{equation}
Here we minimize the group with worst average loss improvement with respect to the previous model $h_1$,  Algorithm \ref{alg:GRM} shows the procedure.

\begin{wrapfigure}[21]{R}{0.5\textwidth}
\begin{minipage}{0.50\textwidth}
\vspace{-2em}
\begin{algorithm}[H]
\scriptsize
    \caption{\scriptsize\textsc{Group-Referenced Learning }}
    \label{alg:GRM}
    {\bfseries Input:} Dataset $D^{tr}=\{x_i,y_i,g_i\}_{i=1}^{n}$, reference model $h_1$, parametric model $h_\theta$,
     $\eta$: model learning rate, $\gamma$: adversary learning rate, batch size $n_B$, aggregation size $N$.
    \begin{algorithmic}[1]
    \State {\bfseries Init parameters and group weights} $\theta^0, \lambda=\{\frac{n_g}{n}\}_{g=1}^{|\mathcal{G}|}$
    \State {\bfseries Compute reference group loss} $\hat{L}_g(h_1)$ using Eq.\ref{eq:GRM}
    \State {\bfseries Expand dataset} $D^{tr}=\{x_i,y_i,g_i, i\}_{i=1}^{n}$ 
    \While {not converged}
    \State {\bfseries Sample aggregated batch} $AB\sim {D^{tr}}^{\otimes (N\times n_B)}$
    \For{$n=1$ {\bfseries to} $N$}
    \State \textbf{Sample batch w/o replacement} $B \sim AB^{\otimes n_b}$
    \State \textbf{Compute group losses in batch}
    \State $n_g(B) = \sum\limits_{i\in B}\mathbf{1}(g_i=g)$
    \State $\hat{L}_g(h_\theta; B) = \frac{1}{n_g(B)} \sum\limits_{i\in B} \mathbf{1}(g_i=g)\ell(h(x_i),y_i)$
    \State \textbf{Update model parameters} 
    \State $\theta \leftarrow \theta + \eta \nabla_\theta(\sum\limits_{g\in \mathcal{G}}\lambda_{g}(\hat{L}_g(h_\theta)-\hat{L}_g(h_1)))$
    \EndFor
    \State \textbf{Update group weights} 
    \State $\lambda \leftarrow \mathop{\Pi}\limits_{\substack{\lambda\ge \epsilon}}(\lambda + \gamma \hat{L}_g(h_\theta; AB)$
    \EndWhile
    \end{algorithmic}
    \Return $h_\theta$
    \end{algorithm} 
\end{minipage}
\end{wrapfigure}

\section{Experiments and results}

We perform experiments on the CIFAR-10, CelebA \cite{liu2018large}, and Waterbird \cite{sagawa2019distributionally} datasets. CIFAR-10 is a standard benchmark for multi-label classification with 10 different classes that we considered as both target and group variables. CelebA contains over 200 thousand facial images with various attribute annotations; similarly to \cite{sagawa2019distributionally}, we use CelebA as a binary classification dataset by predicting the ``blond'' label. We used the Cartesian product of ``blond'' and  ``binary gender'' as the group label,\footnote{Note that we use the ``gender'' binary label as a group descriptor for testing purposes, without attempting to ascribe the broader diversity of gender identities to this limiting notion.} giving 4 groups total. The Waterbird dataset is a subset from the CUB image dataset \cite{WahCUB_200_2011}, where the goal is to predict the type of bird (“waterbird” or “landbird”). As in \cite{sagawa2019distributionally}, we used the Cartesian product of the bird class  and the image background ("land" or "water"), leading to a total of 4 groups. We use ResNet-18 and ResNet-34 architectures \cite{he2016deep} as our base classifiers in all cases, and use SGD with cosine annealing learning rate \cite{loshchilov2016sgdr}. Details are provided in Appendix \ref{sec:experimental_appendix}.

We evaluate the effects that different training and model updating techniques have on NFR in the demographic shift scenario outlined in Section \ref{sec:compatibilitypredefined}. We train a baseline model $h_1$ using either ERM or GMMF on a dataset $D^{1}$ comprising $40\%$ of the original training samples from either Cifar10, CelebA, or Waterbird. To model extreme demographic shift and its effect on NFR, we train a successor model $h_2$ on an expanded dataset $D^{2}$ that comprises all of the samples in $D^{1}$, plus all samples from a given group that were on the original training dataset. Since $h_2$ has access to both an expanded dataset $D^{tr_2}$ and a reference model $h_1$, we evaluate all options for model updating (ERM, GMMF, SRM, GRM).

Experiments include all possible combinations of base training methods (ERM, GMMF), model update algorithms (ERM, GMMF, SRM, GRM), and extension groups (i.e., the experiments are repeated for every possible choice of group extension). Table \ref{table:summary_ermmodelupdate_all_test} summarizes the average performance over these extensions for each dataset, detailed results are available in Appendix \ref{sec:experimental_appendix}. We note that GRM has the smallest negative impact on accuracy, and correspondingly better worst case NFR performance, as expected by design. When transitioning from ERM to GMMF, we see large accuracy improvements on the worst class, but we take a significant accuracy penalty on at least one of the other classes as a result. We note that SRM generally obtains similar results to ERM as an update method, showing that per-sample generalization on test set is a hard-to-achieve objective, which suggests GRM may be a more suitable objective in practice. More work is warranted to tackle this important scenario.

\begin{table}[ht]
\footnotesize
\centering
\scriptsize
\setlength{\tabcolsep}{5pt}
\begin{tabular}{|ll|rrrrrr|rrrrrr|}
\toprule
\multicolumn{2}{|c}{Baseline Method ($h_1$)}& \multicolumn{6}{|c|}{ERM}& \multicolumn{6}{c|}{GMMF}\\
\midrule
  Dataset &  Update       & \multicolumn{2}{c}{$\Delta$Acc} & \multicolumn{2}{c}{1-NFR} & \multicolumn{2}{c}{$\Delta$NFR} &  \multicolumn{2}{c}{$\Delta$Acc} & \multicolumn{2}{c}{1-NFR} & \multicolumn{2}{c|}{$\Delta$NFR}  \\
 &  Method &     min &    max &   min &   max &  min &   max &     min &    max &   min &   max & min &   max \\

\midrule
CelebA      & ERM  & -4.7\% & 7.9\%  & 0.932 & 0.996 & 0.003 & \textbf{0.315} & -48.3\% &  \textbf{5.5\%} & 0.516 & \textbf{1.000} & 0.006 & 0.100\\
        & GRM  & \textbf{-1.5\%} & 24.2\% & \textbf{0.953} & 0.987 & 0.004 & 0.285 & \textbf{-1.7\%}  &  1.3\%  & \textbf{0.958} & 0.987 & 0.037 & 0.074 \\
        &GMMF  & -7.9\% & \textbf{53.1\%} & 0.921 & 0.994 & \textbf{0.006} & 0.090 & -3.4\%  &  1.6\% & 0.953 & 0.983 & \textbf{0.041} & 0.063 \\
        & SRM  & -2.9\% & 6.3\%  & 0.917 & \textbf{0.997} & 0.003 & 0.294 & -41.1\% &  5.2\% & 0.589 & \textbf{1.000} & 0.008 & \textbf{0.102}\\

\midrule

Waterbird   & ERM  & -3.7\% & 7.7\%  & 0.940 & \textbf{0.996} & 0.004 & 0.327 & -15.1\% & \textbf{5.5\%} & 0.833 & \textbf{0.996} & 0.009 & 0.204 \\
            & GRM  & \textbf{-3.6\%} & 10.3\% & \textbf{0.953} & 0.991 & 0.006 & 0.305 & -3.8\%  & 1.9\% & 0.944 & 0.986 & \textbf{0.039} & 0.189 \\
            & GMMF & -5.8\% & \textbf{18.3\%} & 0.931 & \textbf{0.996} & \textbf{0.007} & 0.238 & \textbf{-2.7\%}  & 1.5\% & \textbf{0.946} & 0.985 & 0.038 & 0.188 \\
            & SRM  & -5.3\% & 5.4\%  & 0.932 & 0.995 & 0.005 & \textbf{0.353} & 17.2\%  & 4.0\% & 0.817 & \textbf{0.996} & 0.015 & \textbf{0.209} \\
\midrule

CIFAR-10   & ERM  & -7.7\% & 5.0\% & 0.865 & 0.981 & \textbf{0.040} & \textbf{0.189} & -11.1\% & 5.9\% & 0.842 & 0.981 & 0.030 & \textbf{0.164} \\
            & GRM  & -5.0\% & \textbf{6.9\%} & \textbf{0.902} & 0.986 & 0.035 & 0.178 & \textbf{-5.8\%} & 5.2\% & 0.888 & \textbf{0.986} & 0.030 & 0.159\\
            & GMMF & -6.5\% & 6.0\% & 0.896 & 0.982 & 0.033 & 0.166 & -6.1\% & 5.6\% & \textbf{0.893} & 0.979 & \textbf{0.031} & 0.152\\
            & SRM  & \textbf{4.3\%} & \textbf{6.9\%} & 0.898 & \textbf{0.988} & 0.031 & 0.175 & -24.7\% & \textbf{7.4\%} & 0.714 & 0.979 & 0.024 & 0.139\\
\bottomrule
\end{tabular}
\caption{\footnotesize Minimum and maximum group improvement in accuracy, 1-NFR, and difference w.r.t NFR upper bound ($\Delta$NFR) measured on the test set for each dataset. The initial model (baseline $h_1$) is trained with $40\%$ of the original train partition, and for ERM achieved worst and best group accuracies of $37\%$ and $99\%$ on CelebA, $57\%$ and $99\%$ on Waterbird, and 
$74\%$ and $93\%$ on CIFAR-10. For the GMMF baseline the worst and best group accuracies were $90\%$ and $94\%$ on CelebA, $78\%$ and $94\%$ on Waterbird, and $77\%$ and $95\%$ on CIFAR-10. We report the average over all of the updated dataset $\mathcal{D}^{2}$, where we augmented the original training dataset exclusively with the remaining training samples of one of the groups. We evaluate the effect of each update methodology (ERM, GMMF, GRM, SRM) for both baseline training methods for $h_1$ (ERM or GMMF).}
\label{table:summary_ermmodelupdate_all_test}
\end{table}


\section{Concluding remarks}

In this work we analyzed BC in the context of machine learning and how it can be controlled. We analyzed group BC in the context of iterative model updating when our training dataset is expanded without respecting prior demographic or group proportions. We empirically show how the performance on the groups can be affected when updating ERM or GMMF models. Moreover, we proposed and analyzed two approaches to update the previous model with the goal of minimizing the NFR and the decrease in accuracy of each group in held-out data. Our experimental results on common image classification datasets serve as empirical validation for the ideas advanced in this work. Future work should focus on analyzing the generalization properties of these approaches.

\bibliography{cites}

\bibliographystyle{abbrvnat}

\clearpage

\appendix
\section{Appendix}

\subsection{Model compatibility proof}
\label{sec:proof}

Given the definitions of total compatibility (TC) and negative flip rate (NFR) from Equation \ref{eq:TC}. TC can be further decomposed as the sum of Positive compatibility (PC) and Negative compatibility (NC) as follows
\begin{equation}
\begin{array}{l}
\displaystyle
\text{ $ TC({h}_2,{h}_1|Q) = PC({h}_2,{h}_1|Q) + NC({h}_2,{h}_1|Q)$,} \\
\text{ $  PC({h}_2,{h}_1|Q) =  \mathbb{E}_{ Q}[\mathbf{1}(\bar{h}_1(X)\! =\! Y\wedge\bar{h}_2(X) \!= Y)]$,}\\
\text{ $  NC({h}_2,{h}_1|Q) =  \mathbb{E}_{ Q}[\mathbf{1}(\bar{h}_1(X) \!\not=\! Y\wedge\bar{h}_2(X) \!\not=\! Y)]$,}\\
   \text{ $NFR({h}_1\rightarrow{h}_2| Q) = \mathbb{E}_{{X,Y}\sim Q}[\mathbf{1}(\bar{h}_1(X) = Y\wedge\bar{h}_2(X) \not= Y)]$.}
\end{array}    
\end{equation}

We next proceed to prove Preposition \ref{preposition}, which is extended to include upper bounds on PC and NC that arise naturally from the proof.
\begin{preposition}
Given two models $h_1, h_2$ defined over the same input and output support $\mathcal{X}, \mathcal{Y}$ and a joint distribution $Q_{X,Y}: Supp(Q) \subseteq \mathcal{X}\times \mathcal{Y}$ we have the following:
\begin{equation}
\resizebox{0.85\hsize}{!}{$
\begin{array}{l}
    \displaystyle
    PC({h}_2,{h}_1|Q) \ge Acc(h_1|Q) - \min\big(Acc(h_1|Q) ,Err(h_2|Q)\big), \\
    \displaystyle NC({h}_2,{h}_1|Q) \ge Err(h_1|Q) - \min\big( Err(h_1|Q) ,Acc(h_2|Q)\big), \\
    \displaystyle TC({h}_2,{h}_1|Q) \ge 1 - \min\big(Acc(h_1|Q) ,Err(h_2|Q)\big) - \min\big( Err(h_1|Q) ,Acc(h_2|Q)\big), \\
   \displaystyle NFR(h_1 \rightarrow h_2|Q) \le \min\big(Acc(h_1|Q) ,Err(h_2|Q)\big).
\end{array}$}
\label{eq:lowerboundTCSM}
\end{equation}
\end{preposition}

\begin{proof}
From the definition of NFR, and using the fact that the likelihood of two events occurring simultaneously is smaller than the likelihood of either of the events happening on their own ($P(A\cap B)\le \min (P(A),P(B))$, it follows that
\begin{equation}
\begin{array}{l}
 \displaystyle
   NFR(h_1 \rightarrow h_2|Q) \le \min \big(\mathbb{E}_{{X,Y}\sim Q}[\mathbf{1}(\bar{h}_1(X) = Y)], \mathbb{E}_{{X,Y}\sim Q}[\mathbf{1}(\bar{h}_1(X) \not= Y)]\big), \\
   NFR(h_1 \rightarrow h_2|Q) \le \min\big(Acc(h_1|Q) ,Err(h_2|Q)\big).
   
\end{array}
\label{eq:proof}
\end{equation}

From the definition of PC and the upper bound of NFR it follows that 
\begin{equation}
\begin{array}{l}
 \displaystyle
   PC({h}_2,{h}_1|Q) = Acc(h_1|Q) - NFR(h_1 \rightarrow h_2|Q),\\
   PC({h}_2,{h}_1|Q) \ge Acc(h_1|Q) - \min\big(Acc(h_1|Q) ,Err(h_2|Q)\big).
\end{array}
\label{eq:proof2}
\end{equation}
A symmetrical analysis can be done to obtain the lower bound of NC. Combining the bounds for NC and PC we obtain the lower bound for TC.

\end{proof}

\subsection{Interaction between model compatibility and learning kernel}
\label{sec:compatibilitykernel}

It is common in deep learning to train models to reach near-zero training loss, meaning that models are perfectly compatible on the training dataset. However, due to the overparametrized nature of deep neural networks, the models are unlikely to make identical decisions on a different set of data points. The models may achieve the same average performance on unseen data, but will distribute the errors differently.

We analyze the compatibility between models that are independently obtained  with the same learning algorithm applied to the same dataset $\mathcal{D} \sim P^{\otimes n}_{X,Y}$, $h_1,h_2 \sim P_{h|\mathcal{D}}^{\otimes 2} $. We further consider the binary classification problem where $Y \in \{0,1\}$, $h(x) \in [0,1]$, where the decision $\bar{h}(x)$  is obtained via thresholding, $\bar{h}(x) = \mathbf{1}(h(x) \ge \tau)$, with a predefined threshold $\tau \in [0,1]$. We can express the model output as the sum of the expected model output given the dataset and learning algorithm $\mu_h(X) \triangleq \mathbb{E}_{P_{h|\mathcal{D}}}[h(X)],$ plus an input-dependent error $\epsilon(X)$,
\begin{equation}
\begin{array}{l}
    h_i(x) = \mu_h(x) + \epsilon_i(x), \; \mathbb{E}[\epsilon_i(x)] = 0, \;\forall x \in \mathcal{X}, i = 1,2,
\end{array}
\label{eq:modelaverage}
\end{equation}
with a corresponding decision probability $P(\bar{h}_i(x)=1) = P(\epsilon_i(x) \ge \tau - \mu_h(x)).$

Given an input $X \in \mathcal{X}$, errors $\epsilon_1(X)$ and $\epsilon_2(X)$ are i.i.d., in this setting, TC can be expressed as, \footnote{TC can be expressed as $TC(h_1,h_2| X) = P(h_1(X)\ge\tau)P(h_2(X)\ge\tau) + P(h_1(X)<\tau)P(h_2(X)<\tau) $ since we assume $h_1 \perp h_2|\mathcal{D}$.}
\begin{equation}
\begin{array}{l}
    TC(h_1,h_2| X) = P^2(\epsilon(X) \ge \tau - \mu_h(X)) - 2P(\epsilon(X) \ge \tau - \mu_h(X)) + 1. \\
\end{array}
\label{eq:TCMC}
\end{equation}
For every input $X$, TC reaches its maximum value if $P(\epsilon(X) \ge \tau - \mu_h(X))$ takes value $0$ or $1$. We can identify two ways of accomplishing this, one is choosing a lopsided threshold $\tau$ (e.g., $\tau \simeq 0$ or $\tau \simeq 1$), since $\epsilon(X) +\mu_h(X) \in [0,1] $, this reduces the chances of two models making different decisions, but makes the classifiers non-informative, since the decision is only weakly affected by our learned classifier. The other option is to reduce the variance of the noise variable $\epsilon(X)$; we can use Hoeffding's inequality for this by observing that
\begin{equation}
    P(|\epsilon(X)| \ge |\tau - \mu_h(X)| ) \le \exp[-\frac{(\tau - \mu_h(X))^2}{2\text{Var}[\epsilon(X)]}], 
\end{equation}
which provides an upper bound on the probability of the error being capable of flipping the decision of the current model w.r.t the idealized expected model $\mu_h(X)$. One effective way of addressing noise variance is by augmenting the learning kernel with ensembles, since the variance of the ensemble model decays by a factor of $n^{-2}_e$ w.r.t the model without ensembles, with $n_e$ the number of elements in the ensemble. 

Ensembles can have a major impact on model compatibility, this has been empirically reported in other works such as \cite{yan2021positive}, and this simple analysis provides an explanation for their effectiveness. We empirically showcase the effect ensembles have on several compatibility metrics on CelebA \cite{liu2018large} and CIFAR-10 \cite{cifar}, where we train several independent ResNet networks \cite{he2016deep} (ResNet-18 for Cifar-10 and Resnet-34 for CelebA) on the same training dataset using SGD (effectively taking independent samples from the SGD learning kernel). We used a simple ensemble technique where the average output of $n_e$ independent networks constitute our ensemble model. We show how total compatibility and positive compatibility are effectively managed via ensembling in Table \ref{table:ensembles}. 

\begin{table}[ht]
\scriptsize
\centering
\begin{tabular}{|l|lllll|l|lllll|}
\toprule
& \multicolumn{5}{c|}{CIFAR-10}&&\multicolumn{5}{|c|}{CelebA}\\
\midrule
    model &    Acc &     TC &    $\Delta$TC &     PC &    $\Delta$PC  &model & Acc &     TC &    $\Delta$TC &     PC &    $\Delta$PC \\
\midrule
   base &  0.936 &  0.952 &  0.079 &  0.912 &   0.04  &  base &  0.958 &  0.982 &  0.066 &  0.949 &  0.033\\

   2ens &  0.943 &  0.967 &   0.08 &  0.927 &   0.04 & 2ens &   0.96 &  0.987 &  0.067 &  0.953 &  0.034\\
   5ens &  0.947 &  0.979 &  0.084 &  0.937 &  0.042 & 5ens &  0.961 &  0.992 &   0.07 &  0.957 &  0.035 \\
  10ens &   0.95 &  0.985 &  0.086 &  0.942 &  0.043 & 10ens &   0.96 &  0.994 &  0.073 &  0.957 &  0.037\\
  15ens &   0.95 &  0.989 &  0.088 &  0.945 &  0.044 & 15ens &  0.961 &  0.995 &  0.074 &  0.958 &  0.037\\
  20ens &  0.951 &  0.991 &   0.09 &  0.946 &  0.045 & 20ens &  0.961 &  0.996 &  0.074 &  0.959 &  0.037\\

\bottomrule
\end{tabular}

\caption{Evaluation of ensembles and weight regularization in accuracy (Acc), total compatibility (TC), and positive compatibility (PC) on CelebA and CIFAR-10 datasets. The slackness of PC and TC with respect to their lower bounds in Eq. \ref{eq:lowerboundTC} is also reported as $\Delta$TC and $\Delta$PC respectively. Results are averaged over comparisons between 10 pairs of independent models.}
\label{table:ensembles}
\end{table}

{\color{black}  We observe that the average  accuracy increases for larger ensemble sizes, this effect is well known, and one of the main reasons why ensembles are a popular choice for increasing model performance when feasible. This accuracy increasing effect has diminishing returns with the number of ensembles. Most importantly for our analysis, we note that both total compatibility and positive compatibility also increase significantly as a function of ensemble size, as predicted by the analysis of section \ref{sec:compatibilitykernel}. Interestingly, this increase outpaces the natural growth of TC and PC predicted by the simple increase in accuracy, as shown by the $\Delta$TC and $\Delta$PC columns in the same table, further suggesting that the excess increase is due to the reduction in output variance that ensembles provide as a learning algorithm.  }

\subsubsection*{Confidence informs compatibility}

\begin{figure}[ht] 
\centering
\begin{subfigure}[b]{0.98\textwidth}
\centering
\begin{tabular}{cc}
\includegraphics[bb=0 0 512 512, width=0.45\textwidth]{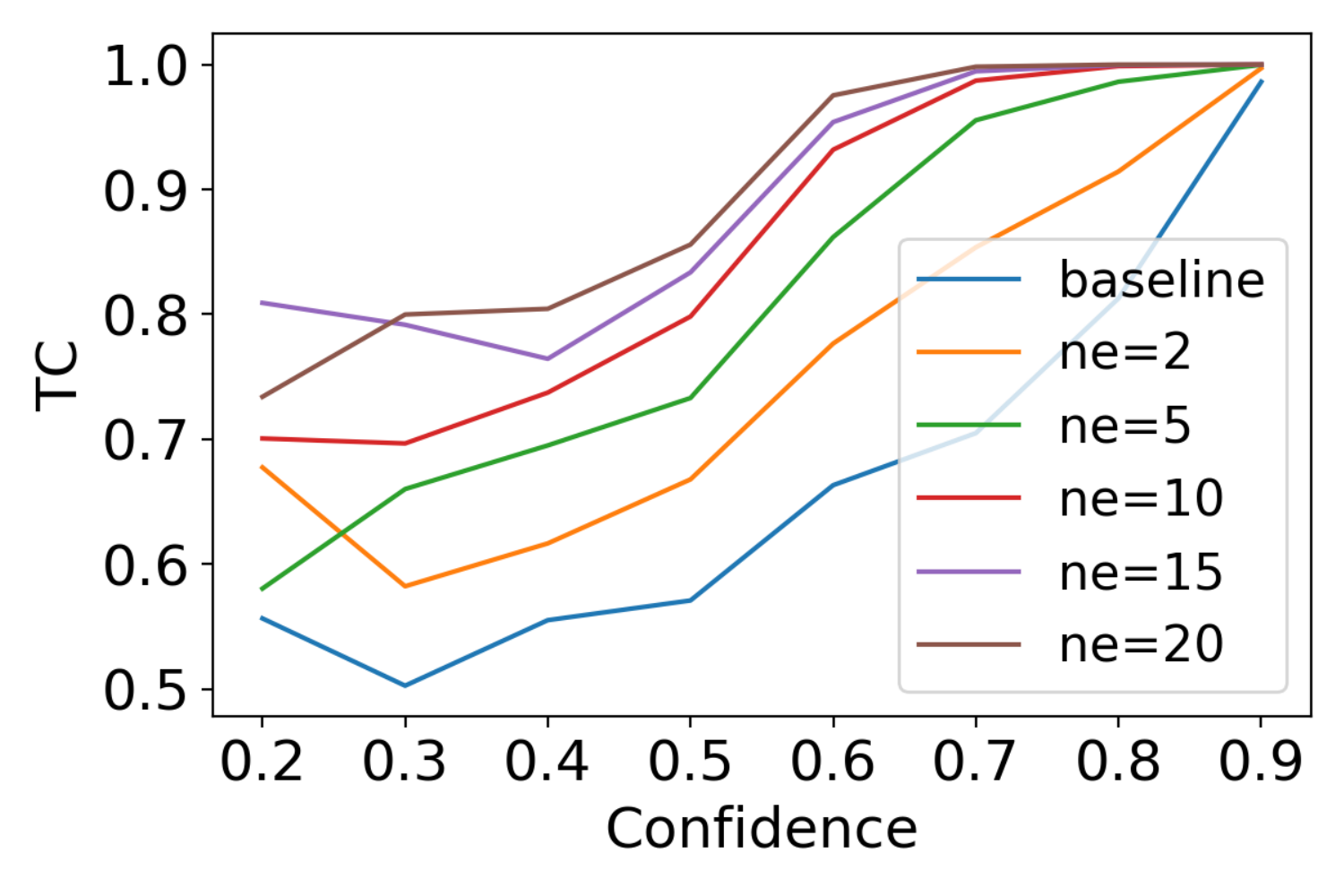} & \includegraphics[bb=0 0 512 512, width=0.45\textwidth]{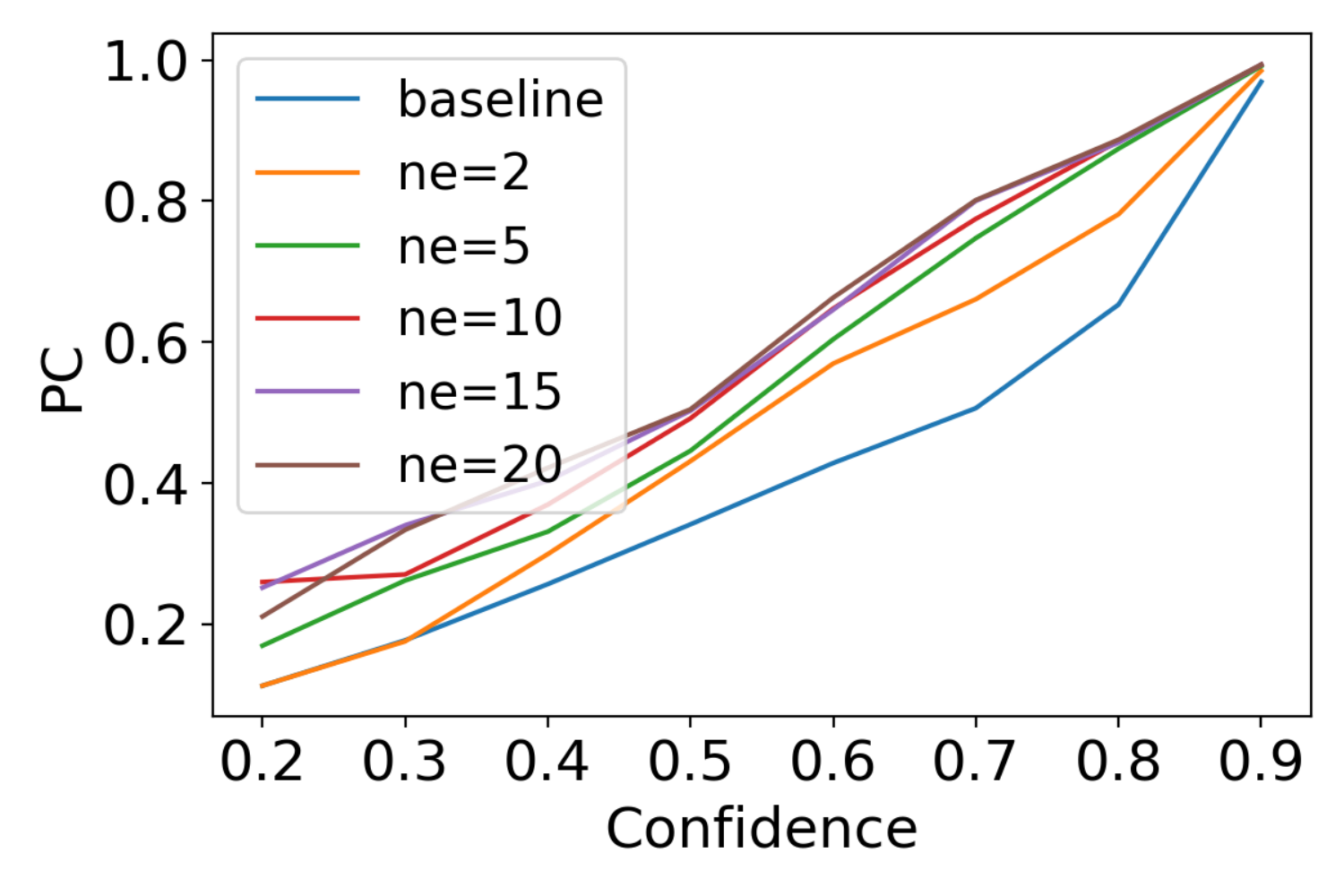}
\end{tabular}
\caption{CIFAR-10}
\end{subfigure}
\begin{subfigure}[b]{0.98\textwidth}
\centering
\begin{tabular}{cc}
\includegraphics[bb=0 0 512 512, width=0.45\textwidth]{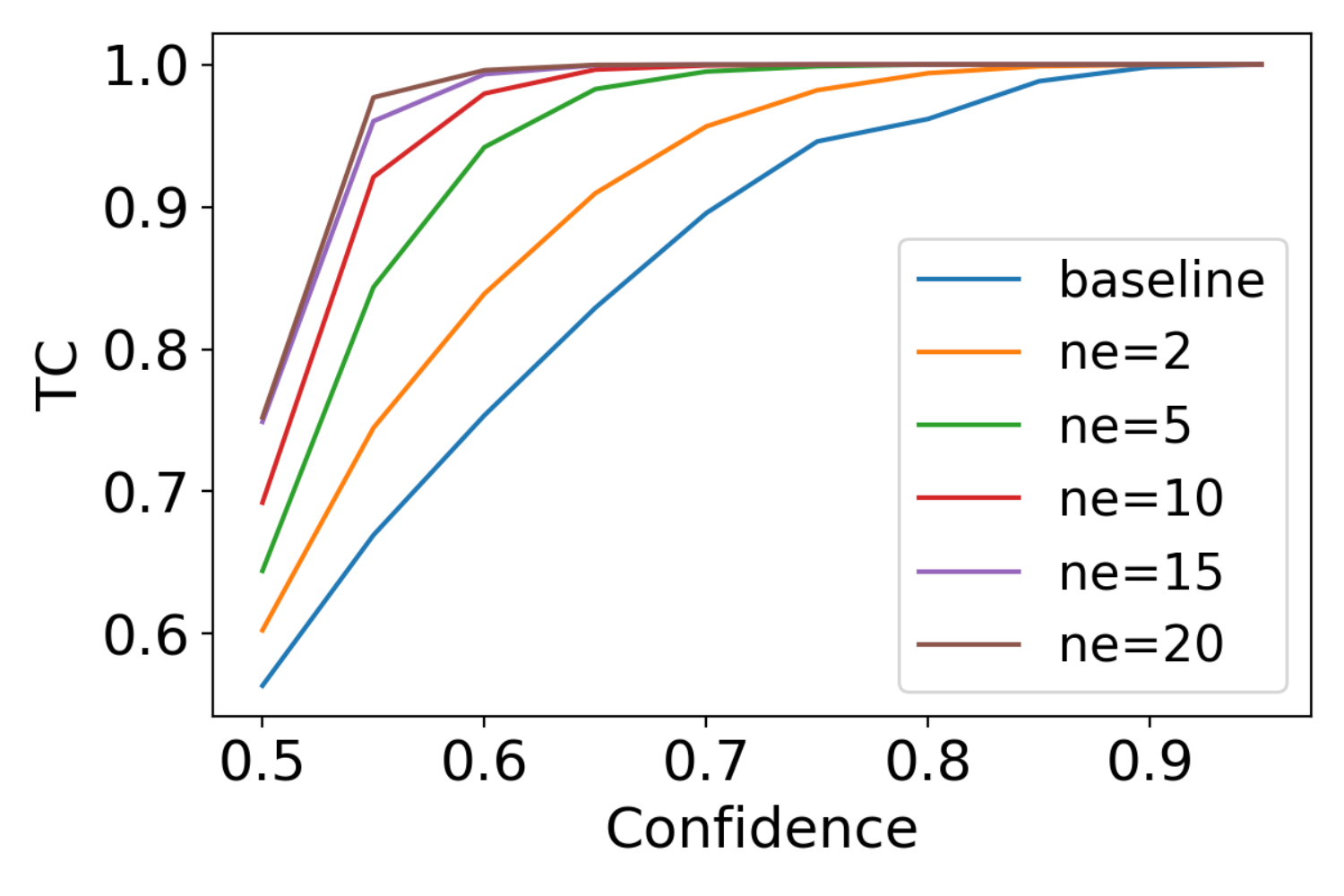} & \includegraphics[bb=0 0 512 512, width=0.45\textwidth]{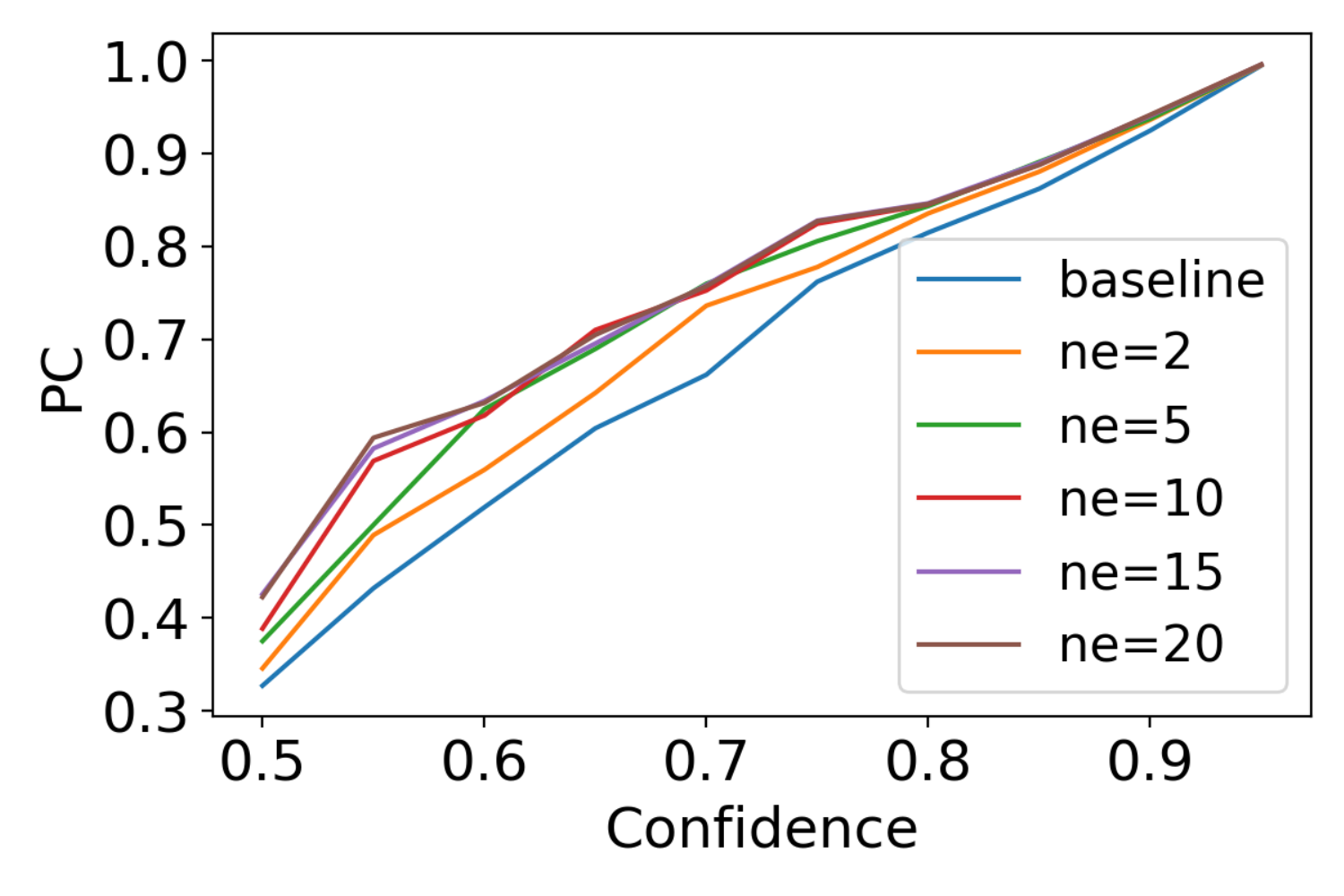}
\end{tabular}
\caption{CelebA}
\end{subfigure}
\caption{ Total compatibility (TC) and positive compatibility (PC) as a function of label confidence on CIFAR-10 and CelebA datasets. Results are shown for varying number of ensembles. We note that confidence is a strong predictor of both TC and PC, and that ensembles uniformly improve on both of these metrics for all confidence values. }
\label{fig:confidence}
\end{figure}

Decision confidence ($\max_{y\in\mathcal{Y}}h^y(x)$) indicates how close a datapoint is to the decision boundary of the model. A large confidence value also indicates robustness to perturbations caused by differences in learned models, since it would require larger $\epsilon(X)$ values to observe different decisions between two models. We verify this observation empirically in the same setting as before. Figure \ref{fig:confidence} shows how TC and PC vary as a function of confidence level, and the improvements introduced by the use of ensembles.

From Figure \ref{fig:confidence} we observe that confidence is indeed a strong predictor of both total and positive compatibility; with lower confidence values associated with a smaller likelihood of samples being compatible across other models. We note that ensembles also share this effect, but the compatibility across models is uniformly improved for all confidence levels.

\subsection{Model learning algorithms}
\label{sec:basemodels_appendix}

Here we provide additional details for all algorithm implementations. Model updating in all cases is done via SGD, so here we provide the empirical objective for all update algorithms and discuss how the adversarial distribution $Q$ is modelled and updated.

To start with, ERM has the following statistical and empirical formulation:
\begin{equation}
    \begin{array}{ll}
       \text{(ERM):}&   \min\limits_{h \in \mathcal{H}} \mathbb{E}_{P}[\ell(h(X),Y)],\\
     & \min\limits_{h \in \mathcal{H}}  \frac{1}{n}\sum_{i=1}^n \ell(h(x_i),y_i).
    \end{array}
\label{eq:ERM_appendix}
\end{equation}

GMMF and GRM can both be likewise expressed in both statistical and empirical formulation as:

\begin{equation}
    \begin{array}{ll}
     \text{(GMMF):}&    \min\limits_{h \in \mathcal{H}} \max\limits_{\substack{ Q_G\ge \epsilon}} \mathbb{E}_{P}[\frac{Q_G}{P_G}\ell(h(X),Y)],\\
     & \min\limits_{h \in \mathcal{H}} \max\limits_{\lambda_g \in \Delta_{\ge\epsilon}^{|\mathcal{G}-1|}} \sum\limits_{g\in \mathcal{G}}\lambda_{g}\sum\limits_{\substack{i=1\\i: g_i=g}}^n \frac{1}{n_{g}}\ell(h(x_i),y_i),
    \end{array}
\label{eq:GMMF_appendix}
\end{equation}

\begin{equation}
    \begin{array}{ll}
        \text{(GRM):}& \min\limits_{h \in \mathcal{H}} \max\limits_{\substack{Q_G \ge \epsilon}}  \mathbb{E}_{P}[\frac{Q_G}{P_G}(\ell(h(X),Y) - L_{G}(h_1))],\\
        & \max\limits_{\lambda_g \in \Delta_{\ge\epsilon}^{|\mathcal{G}-1|}}  \sum\limits_{g\in \mathcal{G}}\lambda_{g}\Big(\sum\limits_{\substack{i=1\\i: g_i=g}}^n \frac{1}{n_{g}}\ell(h(x_i),y_i) - \hat{L}_{g}(h_1)\Big),\\
        & L_{G}(h_1) = \mathbb{E}_{P_{X,Y|G}}[\ell(h_1(X),Y)],\\
        & \hat{L}_{g}(h_1) = \sum\limits_{\substack{i=1\\i: g_i=g}}^n \frac{1}{n_{g}}\ell(h_1(x_i),y_i).
    \end{array}
\label{eq:GRM_appendix}
\end{equation}

Here we use the vector $\lambda_g \in \Delta_{\ge\epsilon}^{|\mathcal{G}-1|} $ to denote the empirical estimate of distributions $Q_G$ satisfying $Q_G\ge \epsilon$, and $n_{g}$ in the empirical formulation represents the number of samples of group $g$. We note that the distribution $Q_G$ represents a potential shift in group composition w.r.t the training distribution $P_G$, the distributions denoted by $P$ indicates the underlying distribution from which the current training dataset is sampled from. The training algorithm for GRM is presented in Algorithm \ref{alg:GRM}, and the algorithm for GMMF is identical if we set all per-group reference values $\hat{L}_g(h_1)=0$.

Finally, SRM is also presented in both formulations:

\begin{equation}
    \begin{array}{ll}

\text{(SRM):}&       \min\limits_{h \in \mathcal{H}} \max\limits_{\substack{Q_{X,Y} }} \mathbb{E}_{ P}[\frac{Q_{X,Y}}{P_{X,Y}}\big( \ell(h(X),Y)-\ell(h_1(X),Y) \big)], \; s.t.: \frac{Q_{X,Y}}{P_{X,Y}}\ge \epsilon\; \forall X,Y,\\
         &\min\limits_{h \in \mathcal{H}} \max\limits_{\lambda \in \Delta_{\ge\epsilon}^{n-1}} \sum_{i=1}^n \frac{\lambda_{i}}{n}(\ell(h(x_i),y_i)-\ell(h_1(x_i),y_i)).
    \end{array}
\label{eq:SRMSM2}
\end{equation}

Here the vector $\lambda \in \Delta_{\ge\epsilon}^{|n-1|} $ already represents the importance weight ratio $\frac{Q_{X,Y}}{P_{X,Y}}$. The algorithm, shown in Algorithm \ref{alg:SRM}, uses PGA on the importance weights $\lambda$.

\begin{algorithm}[H]
    \caption{\footnotesize\textsc{Sample-Referenced Learning }}
    \label{alg:SRM}
    {\bfseries Input:} Dataset $D^{tr}=\{x_i,y_i,g_i\}_{i=1}^{n}$, reference model $h_1$, parametric model $h_\theta$,
     $\eta$: model learning rate, $\gamma$: adversary learning rate, batch size $n_B$, aggregation size $N$.
    \begin{algorithmic}[1]
    \State {\bfseries Init parameters and weights} $\theta^0, \lambda=\{\frac{1}{n}\}_{i=1}^n$ 
    \State {\bfseries Expand dataset}
    \State$\ell_i(h_1)\triangleq\ell(h_1(x_i),y_i)$ 
    \State $D^{tr}=\{x_i,y_i,g_i,\lambda_i,\ell_i(h_1), i\}_{i=1}^{n}$
    \While {not converged}
    \State {\bfseries Sample aggregated batch} $AB\sim {D^{tr}}^{\otimes (N\times n_B)}$
    \For{$n=1$ {\bfseries to} $N$}
\State \textbf{Sample batch w/o replacement} $B \sim AB^{\otimes n_b}$
    \State \textbf{Compute sample losses} $\ell_i(h_\theta) = \ell(h_\theta(x_i),y_i)$
    \State \textbf{Update model parameters} 
    \State $\theta \leftarrow \theta + \eta \nabla_\theta \frac{1}{n_b}\sum_{i\in B}\lambda_i\ell_i(h_\theta)$
    \EndFor
    \State \textbf{Compute weight mass in aggregated batch}
    \State $C =\sum\limits_{i\in AB}\lambda_i $
    \State \textbf{Update weights in aggregated batch} \State $\lambda_{i\in AB} \leftarrow \mathop{\Pi}\limits_{\substack{\lambda\ge \epsilon\\ \sum\limits_{i\in AB}\lambda_i = C}}(\lambda_{i\in AB} + \gamma \ell_{i\in AB}(h_\theta))$
    \EndWhile
    \end{algorithmic}
    \Return $h_\theta$
\end{algorithm}

\subsection{Experimental section extension}
\label{sec:experimental_appendix}

\subsubsection{Architecture and hyperparameters}

Table \ref{tab:hyperparameters} outlines all key architectural choices and hyperarameters used on each dataset.

\begin{table}[]
    \centering
    \scriptsize
    \setlength{\tabcolsep}{5pt}
    \begin{tabular}{|c|ccccccc}
    \toprule
        Dataset & Network & Optimization & Learning rate & Weight decay &Batch size & N & $\eta$    \\
        \midrule
        Waterbirds & ResNet-34 & SGD  & $10^{-3}$ & $10^{-4}$ & 64 & 80 & 0.1\\
        CelebA & ResNet-34 & SGD  & $10^{-3}$ & $10^{-4}$ & 64 & 160 & 0.1\\
        CIFAR-10 & ResNet-18 & SGD  & $10^{-1}$ & $10^{-4}$ & 128 & 80 & 0.1\\
        \bottomrule
    \end{tabular}
    \caption{Summary of architecture and hyperparameters for each dataset. We also use momentum ($0.9$) and cosine annealing}
    \label{tab:hyperparameters}
\end{table}

Other notable changes include reducing the kernel size of the initial convolution layer to $3$ for CIFAR-10 (as standard for this dataset with smaller image sizes). Images for Waterbird and CelebA are scaled, normalized, and cropped to $128\times 128$ for faster training. We use the default train and test partitions for all datasets.

\subsubsection{Additional results}

Here we provide the full experimental results. The initial model (baseline $h_1$) is trained with $40\%$ of the original train partition, and for ERM it achieved worst and best group accuracies of $37\%$ and $99\%$ on CelebA, $57\%$ and $99\%$ on Waterbird, and 
$74\%$ and $93\%$ on CIFAR-10. For the GMMF baseline the worst and best group accuracies were $90\%$ and $94\%$ on CelebA, $78\%$ and $94\%$ on Waterbird, and $77\%$ and $95\%$ on CIFAR-10. We next report results for each updated dataset $\mathcal{D}^{2}$, where we augmented the original training dataset exclusively with the remaining training samples of one of the groups. Tables \ref{table:celebA_ermmodelupdate_all_test}, \ref{table:cub_ermmodelupdate_all_test}, and \ref{table:cifar_ermmodelupdate_all_test} show these results for the CelebA, CIFAR-10, and Waterbirds datasets respectively.

\begin{table}[ht]
\footnotesize
\centering
\setlength{\tabcolsep}{5pt}
\scriptsize
\begin{tabular}{|ll|rrrrrr|rrrrrr|}
\toprule
\multicolumn{2}{|c}{Baseline}& \multicolumn{6}{|c|}{ERM}& \multicolumn{6}{c|}{GMMF}\\
\midrule
  Augmented &  Update       & \multicolumn{2}{c}{$\Delta$Acc} & \multicolumn{2}{c}{1-NFR} & \multicolumn{2}{c}{$\Delta$NFR} &  \multicolumn{2}{c}{$\Delta$Acc} & \multicolumn{2}{c}{1-NFR} & \multicolumn{2}{c|}{$\Delta$NFR}  \\
Group &  Method &     min &    max &   min &   max &  min &   max &     min &    max &   min &   max & min &   max \\
\midrule
non blond & ERM  & -8.6\% & 1.8\%  & 0.900 & 0.998 & 0.003 & 0.288 & -54.8\% &  6.1\% & 0.452 & 1.000 & 0.005 & 0.102\\
non male  & GRM  & -1.9\% & 36.2\% & 0.977 & 0.981 & 0.005 & 0.243 & -2.8\%  &  1.6\%  & 0.949 & 0.991 & 0.036 & 0.079 \\
          & GMMF & -8.0\% & 55.9\% & 0.920 & 0.994 & 0.006 & 0.062 & -3.4\%  &  2.3\% & 0.950 & 0.982 & 0.043 & 0.057\\
          & SRM  & -2.9\% & 4.0\%  & 0.927 & 0.997 & 0.004 & 0.299 & -45.8\% &  5.3\% & 0.542 & 1.000 & 0.008 & 0.102 \\
\midrule
non blond & ERM  & -6.8\% & 0.4\%  & 0.904 & 1.000 & 0.001 & 0.277 & -61.0\% & 5.9\% & 0.390 & 1.000 & 0.002 & 0.102\\
male      & GRM  & -1.9\% & 27.7\% & 0.959 & 0.979 & 0.004 & 0.316 & -1.7\%  &  1.6\%  & 0.955 & 0.991 & 0.036 & 0.073  \\
          & GMMF & -8.0\% & 52.5\% & 0.920 & 0.989 & 0.006 & 0.090 & -3.4\%  & 1.1\% & 0.958 & 0.984 & 0.034 & 0.068 \\
          & SRM  & -4.0\% & 02.4\% & 0.904 & 0.999 & 0.001 & 0.277 & -55.9\% & 5.9\% & 0.441 & 1.000 & 0.002 & 0.102\\
\midrule
blond    & ERM  & -2.6\% & 9.6\%  & 0.966 & 0.996 & 0.003 & 0.339 & -43.5\% &  5.4\% & 0.565 & 0.999 & 0.007 & 0.102 \\
non male & GRM  & -1.7\% & 31.6\% & 0.961 & 0.988 & 0.004 & 0.294 & -1.7\%  &  1.4\%  & 0.960 & 0.988 & 0.034 & 0.079 \\
         & GMMF & -8.1\% & 52.5\% & 0.918 & 0.994 & 0.006 & 0.096 & -3.4\%  &  1.7\% & 0.959 & 0.982 & 0.041 & 0.069\\
         & SRM  & -4.5\% & 1.8\%  & 0.876 & 0.999 & 0.002 & 0.249 & -39.0\% &  5.5\% & 0.610 & 1.000 & 0.006 & 0.102 \\
\midrule
blond    & ERM  & -0.8\% & 19.8\% & 0.958 & 0.991 & 0.004 & 0.356 & -33.9\% &  4.8\% & 0.655 & 0.999 & 0.012 & 0.096 \\
male     & GRM  & -0.4\% & 1.1\%  & 0.915 & 0.998 & 0.003 & 0.288 & -0.5\%  &  0.6\%  & 0.966 & 0.981 & 0.043 & 0.063 \\
         & GMMF & -7.4\% & 51.4\% & 0.926 & 1.000 & 0.006 & 0.113 & -3.2\%  &  1.1\% & 0.944 & 0.985 & 0.047 & 0.056 \\
         & SRM  & -0.3\% & 16.9\% & 0.961 & 0.995 & 0.004 & 0.350 & -23.7\% &  4.4\% & 0.763 & 0.999 & 0.017 & 0.102\\
\midrule
\midrule
Avg     & ERM  & -4.7\% & 7.9\%  & 0.932 & 0.996 & 0.003 & \textbf{0.315} & -48.3\% &  \textbf{5.5\%} & 0.516 & \textbf{1.000} & 0.006 & 0.100\\
        & GRM  & \textbf{-1.5\%} & 24.2\% & \textbf{0.953} & 0.987 & 0.004 & 0.285 & \textbf{-1.7\%}  &  1.3\%  & \textbf{0.958} & 0.987 & 0.037 & 0.074 \\
        &GMMF  & -7.9\% & \textbf{53.1\%} & 0.921 & 0.994 & \textbf{0.006} & 0.090 & -3.4\%  &  1.6\% & 0.953 & 0.983 & \textbf{0.041} & 0.063 \\
        & SRM  & -2.9\% & 6.3\%  & 0.917 & \textbf{0.997} & 0.003 & 0.294 & -41.1\% &  5.2\% & 0.589 & \textbf{1.000} & 0.008 & \textbf{0.102}\\

\bottomrule
\end{tabular}
\caption{\footnotesize CelebA results. Minimum and maximum group improvement in accuracy, 1-NFR, and difference w.r.t NFR upper bound ($\Delta$NFR) in the test split for CelebA dataset. The initial model ($h_1$) is trained with $40\%$ of the original train partition. We evaluate 4 cases for the updated dataset $\mathcal{D}^{2}$, where we augmented the original training dataset exclusively with the remaining training samples of one of the groups (Augmented Group). We evaluate the effect of each update methodology (ERM, GMMF, GRM, SRM) on both of the baseline training methods for $h_1$ (ERM or GMMF).}
\label{table:celebA_ermmodelupdate_all_test}
\end{table}

\begin{table}[ht]
\footnotesize
\centering
\scriptsize
\setlength{\tabcolsep}{5pt}
\begin{tabular}{|ll|rrrrrr|rrrrrr|}
\toprule
\multicolumn{2}{|c}{Baseline}& \multicolumn{6}{|c|}{ERM}& \multicolumn{6}{c|}{GMMF}\\
\midrule
  Augmented &  Update       & \multicolumn{2}{c}{$\Delta$Acc} & \multicolumn{2}{c}{1-NFR} & \multicolumn{2}{c}{$\Delta$NFR} &  \multicolumn{2}{c}{$\Delta$Acc} & \multicolumn{2}{c}{1-NFR} & \multicolumn{2}{c|}{$\Delta$NFR}  \\
Group &  Method &     min &    max &   min &   max &  min &   max &     min &    max &   min &   max & min &   max \\
\midrule
landbird    & ERM  & -5.0\% & 0.3\% & 0.914 & 0.999 & 0.004 & 0.389 & -25.7\% &  5.5\% & 0.740 & 0.999 & 0.003 & 0.217 \\
landback    & GRM  & -4.9\% & 3.0\% & 0.937 & 0.992 & 0.006 & 0.366 & -8.7\% &  2.4\% & 0.900 & 0.991 & 0.026 & 0.207 \\
            & GMMF & -6.4\% & 17.0\% & 0.923 & 0.995 & 0.007 & 0.251 & -3.6\% &  0.1\% & 0.939 & 0.983 & 0.041 & 0.195\\
            & SRM  & -8.4\% & 0.2\% & 0.897 & 0.998 & 0.003 & 0.407 & -27.4\% &  5.2\% & 0.726 & 0.999 & 0.006 & 0.220 \\
\midrule
landbird    & ERM  & -5.0\% & 6.7\% & 0.939 & 0.996 & 0.004 & 0.352 & -16.5\% &  9.0\% & 0.822 & 0.999 & 0.006 & 0.207 \\
waterback   & GRM  & -4.1\% & 17.8\% & 0.958 & 0.995 & 0.006 & 0.243 & -1.3\% &  1.6\% & 0.973 & 0.979 & 0.051 & 0.179 \\
            & GMMF & -5.6\% & 21.0\% & 0.943 & 0.997 & 0.006 & 0.212 & -0.7\% &  2.1\% & 0.969 & 0.980 & 0.047 & 0.178 \\
            & SRM  & -2.3\% & 7.3\% & 0.963 & 0.993 & 0.004 & 0.332 & -13.1\% &  3.1\% & 0.854 & 0.992 & 0.020 & 0.204  \\
\midrule
waterbird   & ERM  & -1.3\% & 15.6\% & 0.975 & 1.000 & 0.006 & 0.269 & -5.1\% &  3.9\% & 0.925 & 0.996 & 0.016 & 0.196\\
landback    & GRM  & -2.0\% & 14.3\% & 0.975 & 0.992 & 0.006 & 0.274 & -1.9\% &  2.2\% & 0.961 & 0.989 & 0.040 & 0.174\\
            & GMMF & -3.9\% & 19.9\% & 0.949 & 0.998 & 0.007 & 0.224 & -1.1\% &  1.6\% & 0.953 & 0.989 & 0.035 & 0.184  \\
            & SRM  & -1.1\% & 9.2\% & 0.969 & 0.991 & 0.006 & 0.324 & -10.1\% &  3.7\% & 0.882 & 0.998 & 0.020 & 0.202 \\
\midrule
waterbird   & ERM  & -3.6\% & 8.3\% & 0.930 & 0.988 & 0.003 & 0.298 & -13.2\% &  3.7\% & 0.844 & 0.988 & 0.010 & 0.196 \\
waterback   & GRM  & -3.3\% & 6.1\% & 0.941 & 0.984 & 0.005 & 0.336 & -3.4\% &  1.4\% & 0.941 & 0.984 & 0.038 & 0.195\\
            & GMMF & -7.4\% & 15.4\% & 0.907 & 0.992 & 0.007 & 0.263 & -5.3\% & 2.2\% & 0.922 & 0.988 & 0.031 & 0.195 \\
            & SRM  & -9.4\% & 4.8\% & 0.900 & 0.998 & 0.006 & 0.350 & -18.2\% &  4.0\% & 0.808 & 0.996 & 0.015 & 0.210 \\
\midrule
\midrule
Avg         & ERM  & -3.7\% & 7.7\%  & 0.940 & \textbf{0.996} & 0.004 & 0.327 & -15.1\% & \textbf{5.5\%} & 0.833 & \textbf{0.996} & 0.009 & 0.204 \\
            & GRM  & \textbf{-3.6\%} & 10.3\% & \textbf{0.953} & 0.991 & 0.006 & 0.305 & -3.8\%  & 1.9\% & 0.944 & 0.986 & \textbf{0.039} & 0.189 \\
            & GMMF & -5.8\% & \textbf{18.3\%} & 0.931 & \textbf{0.996} & \textbf{0.007} & 0.238 & \textbf{-2.7\%}  & 1.5\% & \textbf{0.946} & 0.985 & 0.038 & 0.188 \\
            & SRM  & -5.3\% & 5.4\%  & 0.932 & 0.995 & 0.005 & \textbf{0.353} & 17.2\%  & 4.0\% & 0.817 & \textbf{0.996} & 0.015 & \textbf{0.209} \\
\bottomrule
\end{tabular}
\caption{\footnotesize Waterbird results. Minimum and maximum group improvement in accuracy, 1-NFR, and difference w.r.t NFR upper bound ($\Delta$NFR) in the test split for Waterbird dataset. The initial model ($h_1$) is trained with $40\%$ of the original train partition. We evaluate 4 cases for the updated dataset $\mathcal{D}^{2}$, where we augmented the original training dataset exclusively with the remaining training samples of one of the groups (Augmented Group). We evaluate the effect of each update methodology (ERM, GMMF, GRM, SRM) on both of the baseline training methods for $h_1$ (ERM or GMMF). }
\label{table:cub_ermmodelupdate_all_test}
\end{table}

\begin{table}[ht]
\footnotesize
\centering
\scriptsize
\setlength{\tabcolsep}{5pt}
\begin{tabular}{|ll|rrrrrr|rrrrrr|}
\toprule
\multicolumn{2}{|c}{Baseline} & \multicolumn{6}{|c|}{ERM}& \multicolumn{6}{c|}{GMMF}\\
\midrule
  Augmented & Update       & \multicolumn{2}{c}{$\Delta$Acc} & \multicolumn{2}{c}{1-NFR} & \multicolumn{2}{c}{$\Delta$NFR} & \multicolumn{2}{c}{$\Delta$Acc} & \multicolumn{2}{c}{1-NFR} & \multicolumn{2}{c|}{$\Delta$NFR}  \\
Group & Method &  min & max & min & max & min & max &  min & max & min & max & min & max \\
\midrule
airplane \& & ERM. & -3.9\% & 4.2\% & 0.892 & \text{0.997} & \text{0.032} & \text{0.195} & -5.7\% & 5.4\% & 0.889 & 0.991 & 0.019 & \text{0.168} \\
automobile  & GRM  & -2.5\% & 4.7\% & 0.918 & \text{0.997} & 0.026 & 0.177 &  -2.3\% &  5.8\% & 0.908 & 0.992 & 0.021 & 0.153 \\
            & GMMF & -2.2\% & 5.6\% & 0.922 & \text{0.997} & 0.022 & 0.165 & -2.5\% & 6.0\% & 0.914 & 0.993 & \text{0.020} & 0.161 \\
            & SRM  & \text{-1.8\%} & \text{5.9\%} & \text{0.923} & 0.996 & 0.025 & 0.167 & \text{-1.7\%} & \text{6.5\%} & \text{0.915} & \text{0.996} & 0.016 & 0.154 \\
 \midrule
bird  \&    & ERM  & -9.5\% & 10.1\% & 0.876 & 0.977 & \text{0.044} & \text{0.150} & -8.7\% & 10.6\% & 0.885 & 0.973 & 0.029 & \text{0.136} \\
cat         & GRM  & \text{-4.5\%} & \text{15.0\%} & \text{0.914} & 0.983 & 0.035 & 0.129 & -4.1\% &  5.8\% & 0.916 & 0.985 & 0.030 & 0.133 \\
            & GMMF & -6.8\% & 13.5\% & 0.899 & 0.975 & 0.031 & 0.127 & -3.6\% & 11.1\% & \text{0.918} & 0.977 & 0.036 & 0.118 \\
            & SRM  & -8.0\% & 12.5\% & 0.889 & \text{0.984} & 0.040 & 0.148 & -8.4\% & \text{14.0\%} & 0.883 & \text{0.984} & \text{0.037} & 0.131\\
  \midrule
deer  \&    & ERM  & -8.8\% & 5.5\% & 0.850 & 0.971 & \text{0.047} & 0.202 & -17.4\% & 5.3\% & 0.789 & 0.972 & \text{0.036} & \text{0.185} \\
dog         & GRM  & -7.0\% & \text{7.7\%} & 0.873 & 0.979 & 0.045 & \text{0.207} & -9.7\% &  3.2\% & 0.862 & 0.971 & 0.036 & 0.176 \\
            & GMMF & -13.3\% & 4.3\% & 0.853 & 0.968 & 0.040 & 0.179 & -12.3\% & 5.1\% & 0.857 & 0.972 & 0.032 & 0.156 \\\
            & SRM  & \text{-4.4\%} & 6.8\% & \text{0.885} & \text{0.981} & 0.033 & 0.193 & -10.5\% & \text{8.0\%} & 0.847 & \text{0.991} & 0.032 & 0.174\\
  \midrule
frog  \&    & ERM  & -10.6\% & 2.8\% & 0.834 & 0.976 & \text{0.047} & \text{0.204} & -13.5\% & 4.9\% & 0.813 & 0.979 & 0.037 & 0.170\\
horse       & GRM  & -6.7\% & 4.5\% & \text{0.894} & 0.985 & 0.039 & 0.192 &-9.1\% &  8.3\% & 0.865 & 0.996 & 0.031 & 0.178 \\
            & GMMF & -7.5\% & 4.4\% & 0.884 & 0.984 & 0.040 & 0.189 & \text{-8.6\%} & 1.7\% & \text{0.876} & 0.964 & \text{0.039} & 0.172 \\
            & SRM  & \text{-3.6\%} & \text{5.9\%} & 0.889 & \text{0.991} & 0.031 & 0.183 & -9.2\% & \text{7.2\%} & 0.859 & 0.988 & 0.034 & 0.173 \\
            
 \midrule
ship  \&    & ERM  & -5.9\% & 2.5\% & 0.875 & 0.984 & \text{0.031} & \text{0.195} & -10.1\% & 3.5\% & 0.836 & 0.990 & 0.029 & 0.159 \\
truck       & GRM  & -4.2\% & 2.6\% & 0.911 & 0.987 & 0.029 & 0.184 &  -3.9\% &  3.0\% & 0.890 & 0.986 & 0.034 & 0.157  \\
            & GMMF & \text{-2.8\%} & 2.2\% & \text{0.922} & 0.984 & 0.034 & 0.172 & \text{-3.4\%} & \text{4.0\%} & \text{0.899} & \text{0.991} & 0.030 & 0.155 \\
            & SRM  & -3.8\% & \text{3.4\%} & 0.905 & \text{0.990} & 0.028 & 0.186 & -93.6\% & 1.1\% & 0.064 & 0.936 & 0.000 & 0.064 \\
\midrule
\midrule
Avg         & ERM  & -7.7\% & 5.0\% & 0.865 & 0.981 & \textbf{0.040} & \textbf{0.189} & -11.1\% & 5.9\% & 0.842 & 0.981 & 0.030 & \textbf{0.164} \\
            & GRM  & -5.0\% & \textbf{6.9\%} & \textbf{0.902} & 0.986 & 0.035 & 0.178 & \textbf{-5.8\%} & 5.2\% & 0.888 & \textbf{0.986} & 0.030 & 0.159\\
            & GMMF & -6.5\% & 6.0\% & 0.896 & 0.982 & 0.033 & 0.166 & -6.1\% & 5.6\% & \textbf{0.893} & 0.979 & \textbf{0.031} & 0.152\\
            & SRM  & \textbf{4.3\%} & \textbf{6.9\%} & 0.898 & \textbf{0.988} & 0.031 & 0.175 & -24.7\% & \textbf{7.4\%} & 0.714 & 0.979 & 0.024 & 0.139\\
\bottomrule
\end{tabular}
\caption{\footnotesize CIFAR-10 results.  Minimum and maximum group improvement in accuracy, 1-NFR, and difference w.r.t NFR upper bound ($\Delta$NFR) in the test split for CIFAR-10 dataset. The initial model ($h_1$) is trained with $40\%$ of the original train partition. We evaluate 4 cases for the updated dataset $\mathcal{D}^{2}$, where we augmented the original training dataset exclusively with the remaining training samples of one of the groups (Augmented Group). We evaluate the effect of each update methodology (ERM, GRM, GRM, SRM) on both of the baseline training methods for $h_1$ (ERM or GRM). }
\label{table:cifar_ermmodelupdate_all_test}
\end{table}

\end{document}